\documentclass[journal,twoside,web]{ieeecolor}
\usepackage{generic}
\usepackage{cite}
\usepackage{amsmath,amssymb,amsfonts}
\usepackage{algorithmic}
\usepackage{graphicx}
\usepackage{algorithm,multirow}
\usepackage{hyperref}
\usepackage{soul}
\hypersetup{hidelinks=true}
\usepackage{booktabs}
\usepackage{textcomp}

\def\BibTeX{{\rm B\kern-.05em{\sc i\kern-.025em b}\kern-.08em
    T\kern-.1667em\lower.7ex\hbox{E}\kern-.125emX}}
\begin{document}
\title{Personalized Causal Graph Reasoning for LLMs: An Implementation for Dietary Recommendations}

\author{Zhongqi Yang\textsuperscript{1} and Amir M. Rahmani\textsuperscript{1,2}
\thanks{Department of Computer Science, University of California, Irvine\textsuperscript{1} }
\thanks{School of Nursing, University of California, Irvine\textsuperscript{2} }
\thanks{Submitted to IEEE BHI 2025 in June 2025 }
}


\maketitle

\begin{abstract}
Large Language Models (LLMs) excel at general-purpose reasoning by leveraging broad commonsense knowledge, but they remain limited in tasks requiring personalized reasoning over multifactorial personal data. This limitation constrains their applicability in domains such as healthcare, where decisions must adapt to individual contexts.
We introduce Personalized Causal Graph Reasoning, a framework that enables LLMs to reason over individual-specific causal graphs constructed from longitudinal data. Each graph encodes how user-specific factors influence targeted outcomes. In response to a query, the LLM traverses the graph to identify relevant causal pathways, rank them by estimated impact, simulate potential outcomes, and generate tailored responses.
We implement this framework in the context of nutrient-oriented dietary recommendations, where variability in metabolic responses demands personalized reasoning. Using counterfactual evaluation, we assess the effectiveness of LLM-generated food suggestions for glucose control. Our method reduces postprandial glucose iAUC across three time windows compared to prior approaches. Additional LLM-as-a-judge evaluations further confirm improvements in personalization quality.
\end{abstract}

\begin{IEEEkeywords}
Artificial intelligence, Bioinformatics, Cause effect analysis, Large language models, Precision medicine
\end{IEEEkeywords}

\section{Introduction}
Large Language Models (LLMs) have shown strong performance in general-purpose reasoning tasks, such as instruction following, factual recall, and logical inference~\cite{achiam2023gpt,lan2025large}. These capabilities have motivated interest in applying LLMs to specialized domains like healthcare~\cite{sun2025taming,fu2023kg4nh,yang2024rdguru}, where reasoning involves domain-specific constraints and context-sensitive decisions. 
In such settings, LLMs are expected to handle information that is more structured, variable across individuals, and grounded in expert knowledge.

A particularly compelling use case is nutrition-oriented dietary recommendations, where individual physiological differences play a central role in determining appropriate nutritional guidance. In this setting, LLMs have been explored as conversational agents that suggest meals or food items based on user input~\cite{yang2024chatdiet,wu2025foodcoach,luo2024survey,abbasian2024knowledge}. 
However, current LLMs often fail to adapt to individual metabolic profiles, instead offering generic suggestions that may be suboptimal for specific users.

\begin{figure}[th]
  \includegraphics[width=0.9\columnwidth]{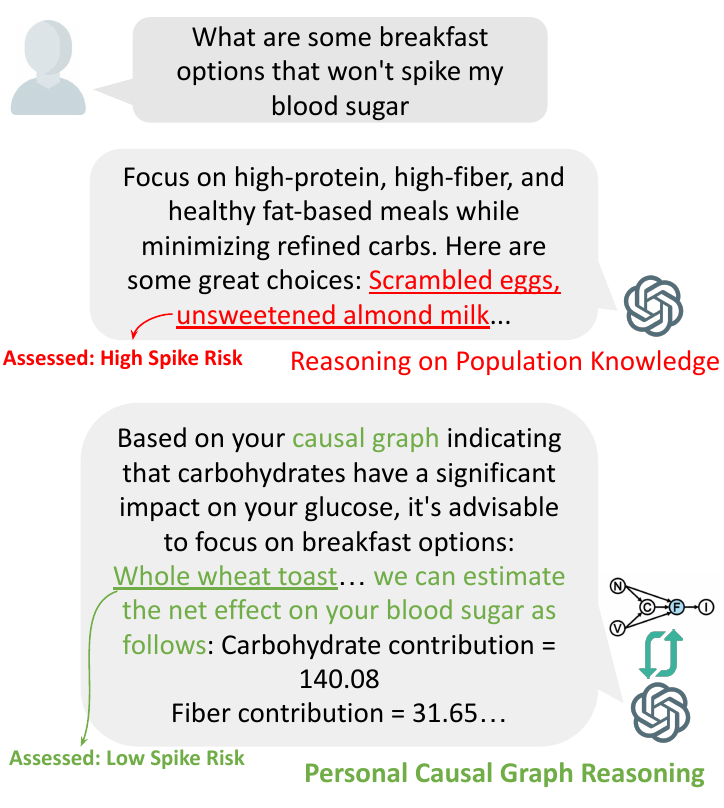}
  \caption{Comparison between a standalone LLM and the proposed Personalized Causal Graph Reasoning for dietary recommendations. The standalone LLM relies on generic reasoning and may provide risky advice, while our method utilizes a personal causal graph to assess individual metabolic responses for more precise recommendations.}
  \label{fig:demo}
\end{figure}

This limitation stems from a fundamental mismatch: human experts rely on a person’s longitudinal data, medical history, and physiological context, while LLMs operate primarily on population-level correlations. Without structured representations of individual-specific health dynamics, LLM-generated recommendations often lack precision and may even be unsafe~\cite{chen2024large,cathy2024physio,halevy2023learnings,tanneru2024hardness,yu2024enhancing,subramanian2024graph}.
Figure~\ref{fig:demo} illustrates this gap. A standard LLM may recommend foods broadly considered healthy, such as almond milk or eggs, without accounting for personal metabolic responses. 
In contrast, an LLM equipped with a personalized causal graph can reason about user-specific glucose sensitivity to nutrients and generate more appropriate, individualized suggestions.

We address this gap by introducing \textbf{Personalized Causal Graph Reasoning}, a framework that augments LLMs with structured, individual-level causal representations.  For each user, we construct a causal graph from longitudinal data that captures how specific inputs influence personal outcomes. 
The LLM traverses the graph in response to the user’s query, identifying causal pathways that connect upstream variables to a user-defined target outcome. It ranks these pathways by estimated causal strength, retrieves relevant candidates from external knowledge sources, simulates potential interventions using the graph structure, and generates personalized responses grounded in the inferred impact of those interventions.
Our method advances upon prior approaches that encode personalization implicitly, enabling explicit and interpretable reasoning grounded in user-specific patterns.

We implement the proposed framework in the context of nutrition-oriented dietary recommendation, where each user’s causal graph captures how specific nutrients influence their glycemic response. To operationalize this, we use real-world data from continuous glucose monitoring (CGM), food intake, and physical activity records to construct personalized causal graphs. The LLM then uses these graphs to generate dietary suggestions aimed at stabilizing blood glucose levels.
To evaluate effectiveness, we introduce a counterfactual simulation method that estimates the expected glycemic impact of recommended foods using each user's personalized causal graph. Additionally, we employ an LLM-as-a-judge to assess the quality of personalized reasoning in the responses.

Our contributions are as follows:
\begin{itemize}
    \item We propose \textbf{Personalized Causal Graph Reasoning}, a general framework that enables LLMs to perform structured, individualized reasoning over personal causal graphs.
    
    \item We implement the framework in the domain of personalized dietary recommendation and introduce a \textbf{counterfactual evaluation method} to estimate the glycemic impact of LLM-generated interventions using individual causal graphs.
\end{itemize}

\section{Related Works}
\subsection{LLMs and Reasoning}
Several techniques were proposed to elevate LLMs' general reasoning tasks.
Chain-of-Thought (CoT) as a classic prompting method enhances problem-solving by enabling the generation of intermediate reasoning steps~\cite{wei2022chain}. 
Building upon CoT, approaches such as Tree of Thoughts (ToT) and Graph of Thoughts (GoT) have been proposed to further refine LLM reasoning in a more structured manner~\cite{yao2024tree,besta2024graph}.
ToT allows models to explore multiple reasoning paths, while GoT models information as an arbitrary graph, to combine various reasoning paths into cohesive outcomes. 
By incorporating structured reasoning techniques, LLMs have demonstrated promising performance in decision-making tasks that require dietary knowledge~\cite{azimi2025evaluation,fu2023kg4nh}.

Beyond prompting techniques, studies have explored iterative reasoning refinements. These include generating multiple reasoning paths and selecting the most consistent one, applying step-wise verification, and integrating feedback mechanisms to improve logical consistency~\cite{havrilla2024glore,li2022making,nathani2023maf}. Additionally, Gao et al. propose meta-reasoning, where LLMs dynamically select and apply different reasoning strategies based on the problem context~\cite{gao2024meta}.
The Reasoning on Graphs (RoG) synergizes LLMs with knowledge graphs to enable faithful and interpretable reasoning~\cite{luo2023reasoning}. 
RoG employs a planning-retrieval-reasoning framework, where relation paths grounded in knowledge graphs are generated as faithful plans. These plans are then used to retrieve valid reasoning paths from the graphs.

Their reliance on large-scale, population-level data limits the applicability in contexts requiring precise, personalized reasoning. This limitation arises because they primarily operate on unstructured text prompts and lack mechanisms to incorporate structured representations of personal information. Consequently, they struggle to model the intricate interplay of personal factors necessary for tailored decision-making. However, approaches like RoG offer promising directions to overcome these challenges by using graphs as a bridge to connect personal data to LLM reasoning.

\subsection{Nutrition-Oriented Diet Recommendations
}
Nutrition-oriented diet recommendation services aim to provide dietary advice tailored to individual health needs. Traditional systems often use collaborative filtering, leveraging user interactions and preferences to generate suggestions~\cite{de2023design,abhari2019systematic,nijman2007method}. However, they fail to capture the complex causal relationships between dietary factors and health outcomes and struggle to adapt dynamically to changes in an individual’s health status~\cite{luo2024survey,verma2018challenges}.

We focus on recent advancements that have explored the performance of LLMs on personalized dietary recommendations~\cite{xue2024dishagent,anjanamma2024personalized,yang2024chatdiet}. 
For instance, ChatDiet combines personal and population models to generate tailored food suggestions~\cite{yang2024chatdiet}.
It employs Retrieval-Augmented Generation (RAG) to retrieve triplets from a pre-constructed causal graph, then structures them into prompts that guide the LLM in recommendation generation. 
While this approach enhances personalization, it relies on a fixed pattern of retrieving specific triplets to inform the LLM's responses. 
Despite its promise, further improvements are needed to enable more structured, adaptive reasoning in LLM-based nutrition systems.

\section{Personalized Causal Graph Reasoning Framework}
\begin{figure}[htb]
  \includegraphics[width=0.99\columnwidth]{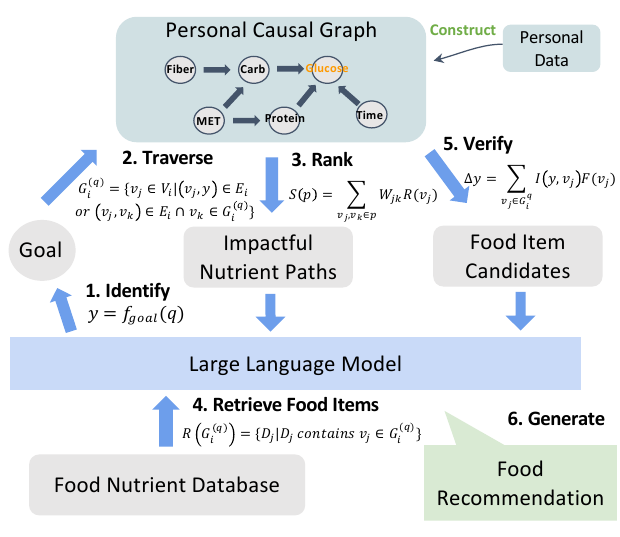}
  \caption{\textbf{Workflow of the Personalized Causal Graph Reasoning.}
Personalized causal graphs are constructed from individual data, enabling an LLM to identify and rank impactful pathways from various factors toward a user-defined objective (e.g., glucose reduction). Based on these impactful pathways, the LLM retrieves candidate items (e.g., food items) from an external knowledge base. These candidate items are then verified by simulating their effects through a counterfactual component. The final output is a personalized recommendation.}
  \label{fig:reason}
\end{figure}

In this section, we introduce the general Personalized Causal Graph Reasoning framework.
The objective of the proposed framework is to enable an LLM agent to reason over a personal causal graph, which encodes the individual's data, such as dietary-health interactions. 
Figure~\ref{fig:reason} illustrates the workflow of this reasoning process on the application of dietary recommendations.
Our proposed framework guides the LLM's reasoning by leveraging the structured causal dependencies between critical factors, such as nutrients, biomarkers, and health outcomes in dietary cases.


We define an individual’s personal causal graph $G_i=(V_i,E_i,W_i)$, where $V_i$ represents the set of nodes (dietary factors, biomarkers, metabolic conditions), $E_i$ denotes the directed causal edges between variables, and $W_i$ encodes the strength of causal relationships.
Given the graph, the LLM agent performs a structured reasoning process that consists of five key stages: goal identification, graph traversal, external knowledge retrieval, verification, and structured response generation.

\subsection{Goal Identification}
When a user submits a query $q$, the LLM first identifies the primary objective and map it to a corresponding node $y$ in the personal causal graph. 
Formally, given the query $q$, the LLM applies a mapping function $f_{goal}$ to determine the target variable:
\begin{equation}
    y=f_{goal}(q)  , \quad y \in V_i
\end{equation}
For instance, if the user asks, \textit{``How can I prevent glucose spikes?"}, the target $y$ would correspond to the glucose incremental area under the curve in the personal causal graph.

\subsection{Personal Causal Graph Traversal and Paths Ranking}
Once the target variable is identified, the LLM agent traverses the personal causal graph to identify relevant dietary factors. The objective is to find upstream nodes (nutrient intake variables) that causally influence $y$. The model retrieves the subgraph $G_i^{(q)}$ consisting of all relevant causal paths leading to $y$:
\begin{equation}
\begin{aligned}
G_i^{(q)} = \{ v_j \in V_i \mid & (v_j, y) \in E_i \text{ or } \\
& (v_j, v_k) \in E_i \land v_k \in G_i^{(q)} \}
\end{aligned}
\end{equation}
The traversal process prioritizes paths based on causal effect strength. Each retrieved path $p=\{v_1,v_2,..., y\}$ is assigned a causal relevance score $S(p)$ computed as:
\begin{equation}
    S(p) = \sum_{(v_j, v_k) \in p} W_{jk} R(v_j)
\end{equation}
where $W_{jk}$ represents the causal strength between 
$v_j$ and $v_k$, and $R(v_j)$ captures the individual’s historical consumption of $v_j$. 
Paths with higher causal scores are given greater weight in generating recommendations. For instance, if an individual has consistently consumed high-glycemic carbohydrates, those pathways might be ranked lower in favor of fiber-rich interventions.

\subsection{External Knowledge Retrieval}
The personal causal graph identifies key dietary factors, but it does not specify which foods to recommend. To bridge this gap, the LLM agent queries a food database from Yang et al.~\cite{yang2024chatdiet} to retrieve relevant nutritional information:
\begin{equation}
    R(G_i^{(q)}) = \{ D_j \mid D_j \text{ contains } v_j \in G_i^{(q)} \}
\end{equation}
where $R(G^{(q)}_i)$ represents the set of foods with a high concentration of impactful nutrients, and $D_j$ denotes an individual food item. 
The retrieved food items are ranked based on their concentration of the identified nutrient.

\subsection{Verification by Simulating Effects}

After selecting a food item and its corresponding nutrient composition, the LLM agent simulates hypothetical dietary interventions using the personalized causal graph. Given the ranked causal paths influencing the user's target outcome $y$, the agent estimates the expected change in $y$ under different nutrient-level adjustments. Specifically, the intervention effect of modifying nutrient intake $v_j$ is computed as:

\begin{equation}
\Delta y = \sum_{v_j \in G_i^{(q)}} I(y, v_j) \cdot F(v_j)
\end{equation}

where $I(y, v_j)$ represents the aggregated causal influence of nutrient $v_j$ on the outcome $y$, and $F(v_j)$ models the user's personalized response function:

\begin{equation}
F(v_j) = \beta_j \cdot \Delta v_j + \epsilon
\end{equation}

Here, $\beta_j$ is the individual response coefficient estimated from historical glucose-nutrient associations, $\Delta v_j$ denotes the proposed change in nutrient intake (e.g., increasing fiber by 15g), and $\epsilon$ captures residual variability with zero-mean expectation.

This simulation step serves as a verification layer: the agent uses the constructed recommendation graph to perform a form of counterfactual analysis, ensuring that the proposed intervention is not only causally plausible but also aligned with the user’s goal. We observed that in some cases, even when the correct nutrient-level causal effects are supplied, the LLM may still fail to generate an appropriate response or achieve the intended goal~\cite{yang2024chatdiet}. 
Motivated by this, this step serves as a sanity-checking step inspired by recent advances in agent-based reasoning frameworks~\cite{chang2024applications}, where the agent reevaluates the predicted impact and explores alternative interventions if the initial recommendation falls short.

\subsection{Response Generation}
To generate the final recommendation, we construct a structured prompt that integrates the retrieved causal graph information, food-nutrient associations, and supporting evidence. The prompt explicitly states the target health outcome $y$, presents the causal pathways derived from $G_i^{(q)}$ in natural language, includes ranked dietary factors based on their relevance, and appends the retrieved food-nutrient content data. 
The LLM is prompted to first explain the causal reasoning before presenting the recommendation to keep responses personalized, interpretable, and grounded in causal inference rather than relying on generic knowledge.



\section{Implementation for Nutrition-Oriented Diet Recommendations}

\subsection{Dataset and Pre-Processing}

We utilize a publicly available dataset comprising 45 participants aged 18 to 69 years with a BMI range of 21–46 kg/m², collected between 2021 and 2024~\cite{gutierrezcgmacros}. 
The cohort includes 15 individuals without diabetes (HbA1c $\textless$ 5.7\%), 16 with prediabetes (5.7\% $\leq$ HbA1c $\leq$ 6.4\%), and 14 with type 2 diabetes (HbA1c $\textgreater$ 6.4\%). 
The dataset spans approximately ten days per participant.
We select data from 34 participants who have complete metabolic equivalent of task (MET) recordings.
For each participant, we select the data of continuous glucose monitoring (CGM) and fitness tracker readings recorded at 1-minute intervals, along with detailed meal records, including total caloric intake and macronutrient composition (carbohydrates, protein, fat, fiber) for each meal (breakfast, lunch, and dinner), and daily average MET.

In this implementation, we define the objective as the \textbf{incremental area under the curve (iAUC) of postprandial glucose levels}.
The iAUC quantifies the body's glycemic response to dietary intake and captures both the magnitude and duration of postprandial glucose excursions~\cite{zeevi2015personalized,reynolds2020wholegrain,floch1990blood}.
We compute the iAUC over three distinct intervals: 30 minutes, 1 hour, and 2 hours following food intake. The 30-minute interval captures the initial glucose rise, which reflects absorption kinetics and early insulin dynamics. The 2-hour interval represents the full postprandial phase and characterizes prolonged glycemic effects and glucose regulation efficiency. The 1-hour interval serves as an intermediate measure to distinguish between transient fluctuations and sustained metabolic responses.
For baseline glucose estimation, conventional approaches define the baseline as the fasting glucose level measured immediately before food intake. This definition may not fully account for individual variability in glycemic patterns. To obtain a more representative baseline, we use the average glucose level over the 24 hour period preceding the meal~\cite{chkroun2023defining}.

\subsection{Personal Causal Graph Construction}
To enable personalized causal graph-based reasoning, we construct a personal causal graph using dietary intake, glucose monitoring, and MET data for each user.
The construction process consists of two key steps: inferring the causal structure using a causal discovery method and estimating causal effects.

\subsubsection{Inferring Causal Structure}\label{sec:discover}
The first step in constructing the causal graph is to determine the structure of causal relationships between dietary factors, metabolic biomarkers, and external modulators. 
We apply the Peter-Clark (PC) algorithm~\cite{spirtes2001causation} to infer a directed acyclic graph that represents the direct causal dependencies between these variables. We use the first half of each user's data for the causal graph construction.

The PC algorithm first detects conditional independence relationships to eliminate non-causal edges, ensuring that only direct dependencies are retained. It then orients the edges by leveraging causal constraints, ensuring that dietary intake variables precede metabolic changes in a physiologically meaningful manner. 
Finally, it adjusts for confounders such as physical activity and baseline glucose levels to prevent spurious associations. The output of this step is a causal graph $G_i=(V_i,E_i)$, where nodes $V_i$ represent dietary intake variables, metabolic biomarkers, and external modulators, while edges $E_i$ encode directed causal relationships between these variables. 
We employ the Causal Discovery Tool library to conduct PC algorithm~\cite{kalainathan2020causal}.
For the PC algorithm, we set the significance level to $\alpha=0.05$ for conditional independence testing. This value controls edge removal and is standard in causal discovery. We used Fisher’s Z test with partial correlations, which is appropriate for continuous biomedical variables with moderate sample sizes. The algorithm was run in parallel across available CPU cores. We also tested $\alpha=0.01$ and $\alpha=0.1$ and found that the overall graph structure remained stable.
At this stage, edges represent structure only and do not yet include quantitative effect strengths, which are estimated in the next step.

\subsubsection{Estimating Causal Effects}\label{sec:inference}
Once the causal structure is identified, we estimate the causal effect strengths of the edges in the graph, quantifying how changes in dietary intake influence metabolic outcomes. We employ Structural Causal Models (SCMs), where each variable is expressed as a function of its direct causes and an independent noise term~\cite{elwert2013graphical}. 
To assign causal effect strengths, we assume a linear SCM that models the impact of each dietary factor on a metabolic outcome as a weighted relationship. 
We then apply regression-based inference to estimate the magnitude of these effects with the first half of each user's data, and use the resulting values as the edge weights $E_i$ in the personalized causal graph $G_i$.

\subsection{Counterfactual Evaluation}

To qualitatively assess whether the recommended food intake contributes to achieving the user’s health goal, we simulate the counterfactual outcome using a reference causal graph and evaluate the recommendation against it. Specifically, we construct a personalized “ground truth” causal graph using the user's full dataset, which serves as an oracle for counterfactual simulation. While true ground-truth causal graphs can only be obtained via interventional studies—often infeasible in nutritional research—this full-data graph approximates the best available estimation of the individual's nutrient-glucose causal structure.

For each user query, the LLM recommends a food item based on the estimated causal graph introduced in Sections~\ref{sec:discover} and~\ref{sec:inference}. To eliminate portion-size bias, all food recommendations are normalized to 500 kcal. We then simulate the expected glucose response by introducing the selected food into the full-data causal graph and computing the predicted impact on the target variable, glucose $iAUC$, via interventional inference:
\begin{equation}
\hat{iAUC} = \mathbb{E}[y \mid do(v_j)]
\end{equation}
where $y$ denotes glucose $iAUC$, $v_j$ is the nutrient vector of the recommended food, and $X_c$ represents relevant confounders such as MET. The counterfactual estimate reflects how the recommended food would affect the user’s glucose response, conditioning on contextual covariates.

To benchmark this effect, we compare $\hat{iAUC}$ to the user’s typical response, $\bar{iAUC}i$, defined as the average glucose response to meals with the user’s historical nutrient composition. The difference is summarized as the \textbf{Mean Glucose Reduction (MGR)}:
\begin{equation}
MGR = \frac{1}{N} \sum{i=1}^{N} (\bar{iAUC}_i - \hat{iAUC}_i)
\end{equation}
A positive MGR indicates that, on average, the recommended foods outperform the user's habitual choices in lowering glucose response.

To simulate realistic evaluation settings, we assume that only a limited portion of the user’s data is available to the LLM during recommendation (e.g., 50\% of the data), while the full graph—constructed using the entire dataset—is reserved for evaluation. This setup reflects real-world personalization, where models must generalize from partial data, while the oracle graph enables robust post-hoc validation. The evaluation thus quantifies how well personalized recommendations inferred from limited data align with optimal responses estimated from the full personal causal structure.

\subsection{Experiment Settings}
To generate personalized food recommendations, we employ GPT-4o as the LLM agent. 
The LLM is instructed to follow the process outlined in Figure~\ref{fig:reason}.
The prompt combines these components: instruction, the user’s query, the retrieved causal paths in a structured format, and the retrieved food-nutrient data.
As we retrieve the most influential causal paths, these paths are then summarized into a natural language description. For instance, if the graph indicates that carbohydrate intake strongly increases postprandial glucose levels, while fiber consumption reduces glucose spikes, the extracted causal summary would be formatted as: \textit{``Carbohydrates have a strong positive causal effect on glucose levels (ranked 1). Fiber has a moderate negative effect, reducing glucose spikes (ranked 2)."}
Additionally, LLM is instructed to first analyze the causal relationships, use the retrieved nutrient information to generate a food recommendation, and then verify the food recommendation.

For testing, we query the agent five times per participant, requesting food recommendations for glucose management across three time windows (30 minutes, 1 hour, and 2 hours), using the baseline glucose levels from the past 2 hours. 
To ensure diversity in recommendations, we impose a constraint preventing the agent from suggesting any food items that were previously recommended in earlier queries for the same participant.

\section{Results}

We compare the proposed method against several baseline models. The baselines include RAG approaches, such as ChatDiet~\cite{yang2024chatdiet} and vanilla RAG models augmented with general dietary guidelines, leveraging either Chain-of-Thought (CoT) prompting or Tree-of-Thought (ToT) reasoning. We also include non-RAG baselines, where a vanilla LLM is tested with and without CoT or ToT prompting. The performance of each method is assessed based on MGR and its standard deviation, as reported in Table~\ref{tab:results1}.


\begin{table*}[ht]
\centering

\begin{tabular}{lcccccc}
\toprule
\multirow{2}{*}{Method} & \multicolumn{2}{c}{30 min MGR} & \multicolumn{2}{c}{1 hr MGR} & \multicolumn{2}{c}{2 hr MGR} \\
\cmidrule(lr){2-3}\cmidrule(lr){4-5}\cmidrule(lr){6-7}
& mean [95\% CI] & $p$ vs Proposed & mean [95\% CI] & $p$ vs Proposed & mean [95\% CI] & $p$ vs Proposed \\
\midrule
Proposed & 19.84 [9.12, 30.56] & n/a & 158.21 [137.40, 179.02] & n/a & 411.56 [385.07, 438.05] & n/a \\
ChatDiet [6] & 33.92 [20.70, 47.14] & $0.089$ & 120.45 [90.11, 150.79] & $0.046$ & 307.12 [264.44, 349.80] & $10^{-4}$ \\
 Guidelines + CoT & 16.38 [-3.69, 36.45] & $0.74$ & -45.72 [-136.47, 45.03] & $ 10^{-6}$ & -79.61 [-153.50, -5.71] & $10^{-8}$ \\
Guidelines + ToT & -18.70 [-45.56, 8.16] & $ 0.002$ & 62.19 [-16.65, 141.03] & $ 0.002 $ & 13.88 [-47.20, 74.96] & $  10^{-6}$ \\
 CoT & -10.59 [-33.61, 12.43] & $ 0.006$ & -49.23 [-120.92, 22.46] & $10^{-6}$ & -64.11 [-156.90, 28.68] & $10^{-8}$ \\
 ToT & 8.77 [-19.46, 37.00] & $ 0.36$ & -6.43 [-66.13, 53.27] & $10^{-4}$ & 63.40 [-23.64, 150.44] & $  10^{-5}$ \\
Sole LLM & 21.40 [3.22, 39.58] & $ 0.81$ & 44.83 [-33.21, 122.87] & $10^{-5}$ & -149.89 [-257.19, -42.59] & $ 10^{-10}$ \\
\bottomrule
\end{tabular}
\caption{MGR and the statistical analysis for baseline models and the proposed Personalized Causal Graph Reasoning framework}
    \label{tab:results1}

\end{table*}

As shown in the results, our approach outperforms the baselines over longer time horizons (1 hour and 2 hours), achieving significantly higher MGR ($p<0.05$) with a lower standard deviation. 
Among the baselines, ChatDiet, a retrieval-augmented model, performs competitively in the short-term window but remains less effective in longer time frames compared to our method. The effect of dietary intake over short durations is inherently variable, making it difficult to determine a significant performance difference. However, over extended time windows, where the physiological impact of food consumption can be estimated with greater confidence, the superior performance of our approach more reliably demonstrates the added value of personalized causal reasoning over static retrieval-based systems.

Models that rely solely on general dietary guidelines or prompting techniques such as CoT and ToT exhibit highly unstable performance, with some configurations even leading to an increase in glucose levels. This instability arises because these models lack access to personalized context, making it impossible to capture an individual’s unique metabolic patterns. These findings reinforce the necessity of explicit causal modeling for effective personalized nutrition recommendations. Overall, our results highlight the crucial role of personalized causal graph reasoning, particularly in dietary interventions. 
Our framework enables the model to generate more effective, stable, and context-aware dietary recommendations tailored to individual metabolic responses.


\subsection{Ablation Study}
We conduct an ablation study by progressively removing key components and evaluating their impact. Specifically, we examine the effect of removing the verification step, disabling the path ranking mechanism, and completely excluding the personal causal graph, thereby testing the model’s performance when relying solely on the LLM. The results are summarized in Table~\ref{tab:ablation}.



\begin{table*}[ht]
\centering

\begin{tabular}{lcccccc}
\toprule
\multirow{2}{*}{Method} & \multicolumn{2}{c}{30 min MGR} & \multicolumn{2}{c}{1 hr MGR} & \multicolumn{2}{c}{2 hr MGR} \\
\cmidrule(lr){2-3}\cmidrule(lr){4-5}\cmidrule(lr){6-7}
& mean [95\% CI] & $p$ vs Full & mean [95\% CI] & $p$ vs Full & mean [95\% CI] & $p$ vs Full \\
\midrule
Full & 19.84 [9.12, 30.56] & n/a & 158.21 [137.40, 179.02] & n/a & 411.56 [385.07, 438.05] & n/a \\
Remove Verification step & 19.16 [8.16, 30.16] & $ 0.87$ & 116.40 [93.91, 138.88] & $0.028$ & 402.74 [373.52, 431.96] & $ 0.31$ \\
Remove Path Ranking & 23.88 [10.48, 37.28] & $ 0.66$ & 95.23 [69.64, 120.83] & $10^{-4}$ & 367.02 [335.93, 398.12] & $ 0.018$ \\
Remove Personal Graph (Sole LLM) & 21.40 [3.22, 39.58] & $ 0.72$ & 44.83 [16.89, 72.76] & $  10^{-6}$ & -149.89 [-257.19, -42.59] & $ 10^{-8}$ \\
\bottomrule
\end{tabular}
\caption{Ablation study results on removing key components.}
    \label{tab:ablation}

\end{table*}

The full model achieves the highest glucose reduction, particularly in the more stable 1-hour and 2-hour time windows.
Removing the verification step results in only a slight decline in performance, indicating that while it is not the primary driver of improvement, it helps refine recommendations in certain corner cases.
In contrast, disabling path ranking leads to a substantial increase in variance, as it plays a core role in prioritizing the most influential nutrients, which is essential for stabilizing glucose impact predictions. Removing the personal causal graph entirely prevents the agent from performing personalized reasoning, rendering the model ineffective at generating meaningful dietary recommendations.

\subsection{Evaluating Reasoning Personalization with LLM-as-a-Judge}
To assess the personalization level of our Personalized Causal Graph Reasoning framework, we employ Llama-3 70B~\cite{dubey2024llama} as an LLM-as-a-judge~\cite{zheng2023judging} to compare its reasoning process against the previous method. 
The evaluation follows a blind comparison setup, where the judge is presented with two outputs in a random order without knowing their source. The judge is instructed to select the response that demonstrates a higher degree of personalization of the reasoning process, considering factors such as whether the reasoning incorporates the user’s unique metabolic patterns, past dietary responses, and personalized causal dependencies; whether the response adapts to the specific health context of the user rather than relying on generic dietary principles; and whether the explanation leverages structured causal insights instead of relying on general nutritional heuristics.

Each comparison is conducted across multiple test cases, and the LLM-as-a-judge selects the more personalized reasoning in each instance. The final win rate reflects the percentage of cases where our model was preferred over ChatDiet. The results, presented in Table~\ref{tab:LLMasajudge}, show that our Personalized Causal Graph Reasoning framework achieves a dominant win rate of 98.43\%.

To validate the reliability of the LLM-as-a-judge evaluation, we additionally conducted a human evaluation on a representative subset of 200 randomly selected test cases.  
Two human evaluators also participated in the study. Both were experienced researchers in machine learning and digital health. They were blinded to model identity and asked to assess responses using three criteria: (1) whether the reasoning incorporated user-specific metabolic patterns and dietary responses, (2) whether the recommendation adapted to the user’s health context rather than generic guidelines, and (3) whether the explanation was grounded in causal dependencies instead of surface-level heuristics. In each assessment, evaluators were blinded to model identity and asked to assess which response better reflected personalized reasoning.
The results from the human evaluators aligned with the LLM-as-a-judge outcomes.
This concordance supports the robustness of using LLMs for comparative personalization assessment and confirms that our method's improvements are perceptible and meaningful to human users.

\begin{table}[htb]
    \centering
    \begin{tabular}{l|c|c}
    \toprule
     & LLM-as-a-Judge & Human Evaluation\\
    \midrule
        \textbf{Proposed} & 98.43\% &86.50\%
   \\
         ChatDiet\cite{yang2024chatdiet}& 1.57\% &13.50\%
  \\

\bottomrule
    \end{tabular}
    \caption{LLM-as-a-Judge and Human Evaluation Win Rates on Reasoning Personalization}
    \label{tab:LLMasajudge}
\end{table}

\section{Discussion}

The findings from this study demonstrate the essential role of structured, personalized causal reasoning in LLMs. 
By explicitly modeling personal causal relationships between dietary factors and physiological responses, our method achieves improvements over conventional LLM approaches. 
This advancement aligns with recent literature that discusses LLM-based clinical recommendations must move beyond generic knowledge toward reasoning grounded in causal and personal data~\cite{zhang2024llm,xu2025kellm,zhang2024causality}.
Our results thus provide practical evidence supporting the integration of causal inference frameworks into LLM-based healthcare applications.

Another insight from our work is the necessity of explicitly structured causal reasoning pathways to ensure stable, accurate predictions of physiological outcomes over extended periods. While simpler retrieval-based systems may perform adequately for short-term responses, longer-term predictions require deeper causal understanding due to reduced inherent variability and clearer physiological patterns~\cite{gendron2024counterfactual}. 
This aligns with established clinical knowledge indicating that metabolic responses to dietary interventions are more predictable over longer timescales~\cite{berry2020personalised,nupponen2015metabolic}. Therefore, future healthcare LLM agents should prioritize causal reasoning capabilities, especially when targeting chronic health conditions where sustained metabolic improvement is critical.
We construct a ground-truth graph for each participant using their full dataset and use it as a reference for evaluation. We note that this graph is not a true clinical ground truth, since glycemic responses can vary across time and context, but rather the most accurate approximation achievable under the available data. It provides a consistent benchmark for simulation-based validation and follows prior causal inference practice where full-data graphs serve as a proxy when interventional data are not feasible. Ultimately, prospective clinical validation, such as randomized controlled trials, is required to establish the effectiveness of the framework in practice.

Furthermore, our ablation study reveals insights that the substantial performance degradation observed upon removing path-ranking highlights the critical role of targeted causal-path prioritization. 
Future research can expand this approach by incorporating uncertainty quantification into path-ranking to further refine recommendation quality and interpretability.

Another contribution is the validation of advanced LLMs as a judge for evaluating personalization quality. 
The agreement between LLM-based and human evaluations underscores the viability of leveraging large foundation models for scalable and cost-effective evaluation of personalized interventions. 
This advance opens new avenues for systematically benchmarking personalized recommendation systems without relying exclusively on resource-intensive human evaluations. Future work could strengthen this approach by developing systematic frameworks and benchmarks specifically designed for LLM-as-a-judge to evaluate personalized reasoning across multiple healthcare domains.

Nevertheless, several areas remain to be addressed to enhance the robustness and clinical applicability of our approach. 
A primary limitation is the model’s current specific focus on glucose management. 
Real-world applications typically require balancing multiple concurrent and potentially conflicting health objectives, including weight management, cardiovascular risk reduction, and micronutrient balance \cite{maillot2009meet}. 
Extending the proposed framework to multi-objective causal reasoning, potentially through Pareto optimization or multi-task learning paradigms, could significantly increase practical utility and clinical relevance.

Additionally, the absence of adaptive stopping criteria during causal graph traversal may limit scalability and efficiency as personalized graphs become more complex. Incorporating adaptive search strategies could improve efficiency by dynamically prioritizing exploration paths based on the current state of knowledge and prediction uncertainty.

Our approach currently relies on relatively limited observational data, leading to uncertainty in causal estimation. Observational datasets inherently suffer from confounding and biases that could limit the reliability of causal effect estimates.
A potential next step is expanding data collection efforts, possibly incorporating more robust observational studies, user-generated dietary logs, and wearable device data spanning longer durations. Such datasets would reduce uncertainty in causal relationships and enhance the precision of recommendations.

Further, the causal graphs constructed in this study omit known physiological confounders that could influence glucose responses. 
Variables such as sleep quality, hormonal fluctuations, stress, and gut microbiome composition have recognized impacts on metabolism but were not explicitly modeled here~\cite{schaafsma2021effect}.
Future studies could significantly strengthen personalized causal modeling by systematically incorporating these additional confounders, either through multimodal data integration or by leveraging existing physiological models as structured priors.

Our current evaluation is primarily based on counterfactual simulations rather than direct clinical validation.
We use it because the dataset does not include external cohorts or prospective outcomes. This restricts validation to within-sample inference. 
Considering real-world outcomes depend heavily on factors such as patient adherence, behavioral adaptation, and environmental influences \cite{fung2019role,zhang2015reinforcement}, we would test the framework on diverse populations or conduct prospective observational and randomized studies with broader resources. For example, prospective randomized controlled trials or long-term observational studies are essential to establish its real-world impact. These potential studies would test whether personalized recommendations improve glucose metrics in controlled settings and sustain benefits in daily life. They would also capture patient adherence, behavioral adaptation, and clinical outcomes that cannot be inferred from counterfactual simulations. Designing such studies requires clear endpoints, long-term follow-up, and workflow integration, but they remain the most reliable way to demonstrate practical effectiveness. Given the dataset constraints, we present these as future directions to strengthen external validity.

Lastly, we believe the proposed framework is promising for clinical use because it enables explicit, interpretable reasoning grounded in graphs, which a sole LLM cannot provide. Such structured personalization aligns with the need for transparent recommendations in healthcare, through which clinicians can understand how specific diet or lifestyle factors affect patients’ outcomes. However, we acknowledge that the translation into practice requires addressing several challenges. For example causal graph construction and LLM reasoning introduce computational demands that may not yet be practical for deployment at scale. The benefit of personalized recommendations depends on patient adherence. In addition, clinical decision-making is inherently complex, and the framework must be designed to complement rather than burden workflows so that clinicians can interpret and act on outputs without loss of efficiency.

\section{Conclusion}
We presented Personalized Causal Graph Reasoning to address the need for personalized LLM reasoning in real-world scenarios. An implementation was conducted by integrating the proposed framework into personalized nutrition-oriented dietary recommendations. 
A counterfactual evaluation method was employed to assess performance without requiring human experts. 
The results showed that the proposed approach improved glucose management compared to retrieval-augmented and prompt-based baselines. 
LLM-as-a-judge results indicated that the proposed method provided more personalized reasoning than existing approaches.

Overall, we have demonstrated the importance of personalized LLM reasoning and the effectiveness of personalized causal graph reasoning in a domain where complex personal data plays a critical role-dietary recommendation. 
A deeper analysis is needed for developing more refined personalized reasoning mechanisms to handle multi-objective decision-making and large-scale personal graph reasoning.
The dietary recommendation study could be extended to incorporate additional confounders and include real-world trials to evaluate its practical effectiveness, which we leave for future work.

\section*{References}
\vspace{-3mm}
\bibliographystyle{IEEEtran}
\bibliography{custom}

\begin{thebibliography}{10}
\providecommand{\url}[1]{#1}
\csname url@samestyle\endcsname
\providecommand{\newblock}{\relax}
\providecommand{\bibinfo}[2]{#2}
\providecommand{\BIBentrySTDinterwordspacing}{\spaceskip=0pt\relax}
\providecommand{\BIBentryALTinterwordstretchfactor}{4}
\providecommand{\BIBentryALTinterwordspacing}{\spaceskip=\fontdimen2\font plus
\BIBentryALTinterwordstretchfactor\fontdimen3\font minus \fontdimen4\font\relax}
\providecommand{\BIBforeignlanguage}[2]{{%
\expandafter\ifx\csname l@#1\endcsname\relax
\typeout{** WARNING: IEEEtran.bst: No hyphenation pattern has been}%
\typeout{** loaded for the language `#1'. Using the pattern for}%
\typeout{** the default language instead.}%
\else
\language=\csname l@#1\endcsname
\fi
#2}}
\providecommand{\BIBdecl}{\relax}
\BIBdecl

\bibitem{achiam2023gpt}
J.~Achiam, S.~Adler, S.~Agarwal, L.~Ahmad, I.~Akkaya, F.~L. Aleman, D.~Almeida, J.~Altenschmidt, S.~Altman, S.~Anadkat \emph{et~al.}, ``Gpt-4 technical report,'' \emph{arXiv preprint arXiv:2303.08774}, 2023.

\bibitem{lan2025large}
W.~Lan, Z.~Tang, M.~Liu, Q.~Chen, W.~Peng, Y.~P. Chen, and Y.~Pan, ``The large language models on biomedical data analysis: A survey,'' \emph{IEEE Journal of Biomedical and Health Informatics}, 2025.

\bibitem{sun2025taming}
L.~Sun, D.~Liu, M.~Wang, Y.~Han, Y.~Zhang, B.~Zhou, Y.~Ren \emph{et~al.}, ``Taming unleashed large language models with blockchain for massive personalized reliable healthcare,'' \emph{IEEE Journal of Biomedical and Health Informatics}, 2025.

\bibitem{fu2023kg4nh}
C.~Fu, X.~Pan, J.~Wu, J.~Cai, Z.~Huang, F.~van Harmelen, W.~Zhao, X.~Jiang, and T.~He, ``Kg4nh: a comprehensive knowledge graph for question answering in dietary nutrition and human health,'' \emph{IEEE journal of biomedical and health informatics}, 2023.

\bibitem{yang2024rdguru}
J.~Yang, L.~Shu, H.~Duan, and H.~Li, ``Rdguru: a conversational intelligent agent for rare diseases,'' \emph{IEEE Journal of Biomedical and Health Informatics}, 2024.

\bibitem{yang2024chatdiet}
Z.~Yang, E.~Khatibi, N.~Nagesh, M.~Abbasian, I.~Azimi, R.~Jain, and A.~M. Rahmani, ``Chatdiet: Empowering personalized nutrition-oriented food recommender chatbots through an llm-augmented framework,'' \emph{Smart Health}, vol.~32, p. 100465, 2024.

\bibitem{wu2025foodcoach}
J.~Wu, S.~Mayer, S.~Pilz, Y.~S. Antille, J.~L. Albert, M.~Stoll, K.~Garcia, K.~Fuchs, L.~Bally, L.~Eichelberger \emph{et~al.}, ``Foodcoach: Fully automated diet counseling,'' \emph{IEEE Journal of Biomedical and Health Informatics}, 2025.

\bibitem{luo2024survey}
H.~Luo, F.~Zhuang, R.~Xie, H.~Zhu, D.~Wang, Z.~An, and Y.~Xu, ``A survey on causal inference for recommendation,'' \emph{The Innovation}, 2024.

\bibitem{abbasian2024knowledge}
M.~Abbasian, Z.~Yang, E.~Khatibi, P.~Zhang, N.~Nagesh, I.~Azimi, R.~Jain, and A.~M. Rahmani, ``Knowledge-infused llm-powered conversational health agent: A case study for diabetes patients,'' in \emph{2024 46th Annual International Conference of the IEEE Engineering in Medicine and Biology Society (EMBC)}.\hskip 1em plus 0.5em minus 0.4em\relax IEEE, 2024, pp. 1--4.

\bibitem{chen2024large}
J.~Chen, Z.~Liu, X.~Huang, C.~Wu, Q.~Liu, G.~Jiang, Y.~Pu, Y.~Lei, X.~Chen, X.~Wang \emph{et~al.}, ``When large language models meet personalization: Perspectives of challenges and opportunities,'' \emph{World Wide Web}, vol.~27, no.~4, p.~42, 2024.

\bibitem{cathy2024physio}
C.~M. Fang, V.~Danry, N.~Whitmore, A.~Bao, A.~Hutchison, C.~Pierce, and P.~Maes, ``Physiollm: Supporting personalized health insights with wearables and large language models,'' in \emph{2024 IEEE EMBS International Conference on Biomedical and Health Informatics (BHI)}, 2024, pp. 1--8.

\bibitem{halevy2023learnings}
A.~Halevy and J.~Dwivedi-Yu, ``Learnings from data integration for augmented language models,'' \emph{arXiv preprint arXiv:2304.04576}, 2023.

\bibitem{tanneru2024hardness}
S.~H. Tanneru, D.~Ley, C.~Agarwal, and H.~Lakkaraju, ``On the hardness of faithful chain-of-thought reasoning in large language models,'' \emph{arXiv preprint arXiv:2406.10625}, 2024.

\bibitem{yu2024enhancing}
H.~Yu, C.~Yu, Z.~Wang, D.~Zou, and H.~Qin, ``Enhancing healthcare through large language models: A study on medical question answering,'' in \emph{2024 IEEE 6th International Conference on Power, Intelligent Computing and Systems (ICPICS)}.\hskip 1em plus 0.5em minus 0.4em\relax IEEE, 2024, pp. 895--900.

\bibitem{subramanian2024graph}
A.~Subramanian, Z.~Yang, I.~Azimi, and A.~M. Rahmani, ``Graph-augmented llms for personalized health insights: A case study in sleep analysis,'' in \emph{2024 IEEE 20th International Conference on Body Sensor Networks (BSN)}.\hskip 1em plus 0.5em minus 0.4em\relax IEEE, 2024, pp. 1--4.

\bibitem{wei2022chain}
J.~Wei, X.~Wang, D.~Schuurmans, M.~Bosma, F.~Xia, E.~Chi, Q.~V. Le, D.~Zhou \emph{et~al.}, ``Chain-of-thought prompting elicits reasoning in large language models,'' \emph{Advances in neural information processing systems}, vol.~35, pp. 24\,824--24\,837, 2022.

\bibitem{yao2024tree}
S.~Yao, D.~Yu, J.~Zhao, I.~Shafran, T.~Griffiths, Y.~Cao, and K.~Narasimhan, ``Tree of thoughts: Deliberate problem solving with large language models,'' \emph{Advances in Neural Information Processing Systems}, vol.~36, 2024.

\bibitem{besta2024graph}
M.~Besta, N.~Blach, A.~Kubicek, R.~Gerstenberger, M.~Podstawski, L.~Gianinazzi, J.~Gajda, T.~Lehmann, H.~Niewiadomski, P.~Nyczyk \emph{et~al.}, ``Graph of thoughts: Solving elaborate problems with large language models,'' in \emph{Proceedings of the AAAI Conference on Artificial Intelligence}, vol.~38, 2024, pp. 17\,682--17\,690.

\bibitem{azimi2025evaluation}
I.~Azimi, M.~Qi, L.~Wang, A.~M. Rahmani, and Y.~Li, ``Evaluation of llms accuracy and consistency in the registered dietitian exam through prompt engineering and knowledge retrieval,'' \emph{Scientific Reports}, vol.~15, no.~1, p. 1506, 2025.

\bibitem{havrilla2024glore}
A.~Havrilla, S.~Raparthy, C.~Nalmpantis, J.~Dwivedi-Yu, M.~Zhuravinskyi, E.~Hambro, and R.~Raileanu, ``Glore: When, where, and how to improve llm reasoning via global and local refinements,'' \emph{arXiv preprint arXiv:2402.10963}, 2024.

\bibitem{li2022making}
Y.~Li, Z.~Lin, S.~Zhang, Q.~Fu, B.~Chen, J.-G. Lou, and W.~Chen, ``Making large language models better reasoners with step-aware verifier,'' \emph{arXiv preprint arXiv:2206.02336}, 2022.

\bibitem{nathani2023maf}
D.~Nathani, D.~Wang, L.~Pan, and W.~Y. Wang, ``Maf: Multi-aspect feedback for improving reasoning in large language models,'' \emph{arXiv preprint arXiv:2310.12426}, 2023.

\bibitem{gao2024meta}
P.~Gao, A.~Xie, S.~Mao, W.~Wu, Y.~Xia, H.~Mi, and F.~Wei, ``Meta reasoning for large language models,'' \emph{arXiv preprint arXiv:2406.11698}, 2024.

\bibitem{luo2023reasoning}
L.~Luo, Y.-F. Li, G.~Haffari, and S.~Pan, ``Reasoning on graphs: Faithful and interpretable large language model reasoning,'' \emph{arXiv preprint arXiv:2310.01061}, 2023.

\bibitem{de2023design}
I.~M. de~Hoogh, M.~J. Reinders, E.~L. Doets, F.~P. Hoevenaars, and J.~L. Top, ``Design issues in personalized nutrition advice systems,'' \emph{Journal of Medical Internet Research}, vol.~25, p. e37667, 2023.

\bibitem{abhari2019systematic}
S.~Abhari, R.~Safdari, L.~Azadbakht, K.~B. Lankarani, S.~R.~N. Kalhori, B.~Honarvar, K.~Abhari, S.~Ayyoubzadeh, Z.~Karbasi, S.~Zakerabasali \emph{et~al.}, ``A systematic review of nutrition recommendation systems: with focus on technical aspects,'' \emph{Journal of biomedical physics \& engineering}, vol.~9, no.~6, p. 591, 2019.

\bibitem{nijman2007method}
C.~Nijman, I.~Zijp, A.~Sierksma, A.~Roodenburg, R.~Leenen, C.~Van~den Kerkhoff, J.~Weststrate, and G.~Meijer, ``A method to improve the nutritional quality of foods and beverages based on dietary recommendations,'' \emph{European Journal of Clinical Nutrition}, vol.~61, no.~4, pp. 461--471, 2007.

\bibitem{verma2018challenges}
M.~Verma, R.~Hontecillas, N.~Tubau-Juni, V.~Abedi, and J.~Bassaganya-Riera, ``Challenges in personalized nutrition and health,'' p. 117, 2018.

\bibitem{xue2024dishagent}
C.~Xue, Y.~Guo, Z.~Wang, M.~Shimizu, J.~Jeung, and H.~Mi, ``Dishagent: Enhancing dining experiences through llm-based smart dishes,'' in \emph{Adjunct Proceedings of the 37th Annual ACM Symposium on User Interface Software and Technology}, 2024, pp. 1--4.

\bibitem{anjanamma2024personalized}
C.~Anjanamma, G.~Sirisha, B.~Sravani, K.~Shilpa, C.~L. Narayana, and V.~Vivekanandhan, ``Personalized food nutrient recommendations for kids using ai and behavior analysis,'' in \emph{2024 9th International Conference on Communication and Electronics Systems (ICCES)}.\hskip 1em plus 0.5em minus 0.4em\relax IEEE, 2024, pp. 1805--1810.

\bibitem{chang2024applications}
Y.~Chang, J.-m. Yin, J.-m. Li, C.~Liu, L.-y. Cao, and S.-y. Lin, ``Applications and future prospects of medical llms: A survey based on the m-kat conceptual framework,'' \emph{Journal of Medical Systems}, vol.~48, no.~1, pp. 1--18, 2024.

\bibitem{gutierrezcgmacros}
R.~Gutierrez-Osuna, D.~Kerr, B.~Mortazavi, and A.~Das, ``Cgmacros: a scientific dataset for personalized nutrition and diet monitoring,'' \emph{Scientific Data (under review)}, 2025.

\bibitem{zeevi2015personalized}
D.~Zeevi, T.~Korem, N.~Zmora, D.~Israeli, D.~Rothschild, A.~Weinberger, O.~Ben-Yacov, D.~Lador, T.~Avnit-Sagi, M.~Lotan-Pompan \emph{et~al.}, ``Personalized nutrition by prediction of glycemic responses,'' \emph{Cell}, vol. 163, no.~5, pp. 1079--1094, 2015.

\bibitem{reynolds2020wholegrain}
A.~N. Reynolds, J.~Mann, M.~Elbalshy, E.~Mete, C.~Robinson, I.~Oey, P.~Silcock, N.~Downes, T.~Perry, and L.~Te~Morenga, ``Wholegrain particle size influences postprandial glycemia in type 2 diabetes: a randomized crossover study comparing four wholegrain breads,'' \emph{Diabetes care}, vol.~43, no.~2, pp. 476--479, 2020.

\bibitem{floch1990blood}
J.-P.~L. Floch, P.~Escuyer, E.~Baudin, D.~Baudon, and L.~Perlemuter, ``Blood glucose area under the curve: methodological aspects,'' \emph{Diabetes care}, vol.~13, no.~2, pp. 172--175, 1990.

\bibitem{chkroun2023defining}
C.~Chkroun, I.~Trouwborst, A.~Cherta-Murillo, L.~Owen, C.~Darimont, and A.~Rytz, ``Defining a continuous glucose baseline to assess the impact of nutritional interventions,'' \emph{Frontiers in Nutrition}, vol.~10, p. 1203899, 2023.

\bibitem{spirtes2001causation}
P.~Spirtes, C.~Glymour, and R.~Scheines, \emph{Causation, prediction, and search}.\hskip 1em plus 0.5em minus 0.4em\relax MIT press, 2001.

\bibitem{kalainathan2020causal}
D.~Kalainathan, O.~Goudet, and R.~Dutta, ``Causal discovery toolbox: Uncovering causal relationships in python,'' \emph{Journal of Machine Learning Research}, vol.~21, no.~37, pp. 1--5, 2020.

\bibitem{elwert2013graphical}
F.~Elwert, ``Graphical causal models,'' in \emph{Handbook of causal analysis for social research}.\hskip 1em plus 0.5em minus 0.4em\relax Springer, 2013, pp. 245--273.

\bibitem{dubey2024llama}
A.~Dubey, A.~Jauhri, A.~Pandey, A.~Kadian, A.~Al-Dahle, A.~Letman, A.~Mathur, A.~Schelten, A.~Yang, A.~Fan \emph{et~al.}, ``The llama 3 herd of models,'' \emph{arXiv preprint arXiv:2407.21783}, 2024.

\bibitem{zheng2023judging}
L.~Zheng, W.-L. Chiang, Y.~Sheng, S.~Zhuang, Z.~Wu, Y.~Zhuang, Z.~Lin, Z.~Li, D.~Li, E.~Xing \emph{et~al.}, ``Judging llm-as-a-judge with mt-bench and chatbot arena,'' \emph{Advances in Neural Information Processing Systems}, vol.~36, pp. 46\,595--46\,623, 2023.

\bibitem{zhang2024llm}
Y.~Zhang, H.~Liu, Y.~Xiao, M.~Amoon, D.~Zhang, D.~Wang, S.~Yang, and C.~Quek, ``Llm-enhanced multi-teacher knowledge distillation for modality-incomplete emotion recognition in daily healthcare,'' \emph{IEEE Journal of Biomedical and Health Informatics}, 2024.

\bibitem{xu2025kellm}
T.~Xu and B.~Li, ``Kellm: Knowledge-enhanced label-wise large language model for safe and interpretable drug recommendation.'' \emph{Electronics (2079-9292)}, vol.~14, no.~1, 2025.

\bibitem{zhang2024causality}
Y.~Zhang, J.~You, Y.~Bai, J.~Zhang, K.~Bao, W.~Wang, and T.-S. Chua, ``Causality-enhanced behavior sequence modeling in llms for personalized recommendation,'' \emph{arXiv preprint arXiv:2410.22809}, 2024.

\bibitem{gendron2024counterfactual}
G.~Gendron, J.~M. Ro{\v{z}}anec, M.~Witbrock, and G.~Dobbie, ``Counterfactual causal inference in natural language with large language models,'' \emph{arXiv preprint arXiv:2410.06392}, 2024.

\bibitem{berry2020personalised}
S.~Berry, D.~Drew, I.~Linenberg, J.~Wolf, G.~Hadjigeorgiou, R.~Davies, H.~Al~Khatib, D.~Hart, G.~Surdulescu, D.~Yarand \emph{et~al.}, ``Personalised responses to dietary composition trial (predict): an intervention study to determine inter-individual differences in postprandial response to foods,'' \emph{Research Square}, 2020.

\bibitem{nupponen2015metabolic}
M.~Nupponen, K.~Pahkala, M.~Juonala, C.~G. Magnussen, H.~Niinikoski, T.~R{\"o}nnemaa, J.~S. Viikari, M.~Saarinen, H.~Lagstr{\"o}m, A.~Jula \emph{et~al.}, ``Metabolic syndrome from adolescence to early adulthood: effect of infancy-onset dietary counseling of low saturated fat: the special turku coronary risk factor intervention project (strip),'' \emph{Circulation}, vol. 131, no.~7, pp. 605--613, 2015.

\bibitem{maillot2009meet}
M.~Maillot, F.~Vieux, E.~F. Ferguson, J.-L. Volatier, M.~J. Amiot, and N.~Darmon, ``To meet nutrient recommendations, most french adults need to expand their habitual food repertoire,'' \emph{The Journal of nutrition}, vol. 139, no.~9, pp. 1721--1727, 2009.

\bibitem{schaafsma2021effect}
A.~Schaafsma, L.~Mallee, M.~van~den Belt, E.~Floris, G.~Kortman, J.~Veldman, D.~van~den Ende, and A.~Kardinaal, ``The effect of a whey-protein and galacto-oligosaccharides based product on parameters of sleep quality, stress, and gut microbiota in apparently healthy adults with moderate sleep disturbances: a randomized controlled cross-over study,'' \emph{Nutrients}, vol.~13, no.~7, p. 2204, 2021.

\bibitem{fung2019role}
T.~K. Fung, ``The role of counterfactual thinking in narrative persuasion: Its impact on patients’ adherence to treatment regimen,'' \emph{Health Communication}, 2019.

\bibitem{zhang2015reinforcement}
Y.~Zhang, J.~Paik, and P.~Pirolli, ``Reinforcement learning and counterfactual reasoning explain adaptive behavior in a changing environment,'' \emph{Topics in Cognitive Science}, vol.~7, no.~2, pp. 368--381, 2015.

\end{thebibliography}

\newpage
\section*{Appendix}

\subsection*{Implementation of the Causal Summary}
During recommendation generation, the LLM is instructed to first summarize the most influential causal paths from the personal graph. 
This step ensures that downstream reasoning is grounded in the personalized causal structure rather than relying on generic dietary heuristics. 
We construct the causal summary by ranking the nutrient-to-glucose paths by their estimated causal strength and then presenting them to the LLM in structured text.

\paragraph{System Prompt.}
\begin{quote}
You are given a personal causal graph that links nutrients to glucose outcomes. 
Summarize the top three causal relationships in plain language, ranked by strength. 
Clearly state whether each nutrient increases or decreases glucose response. 
Do not make recommendations yet; only provide a structured causal summary.
\end{quote}

\paragraph{Input Example (from causal paths).}
\begin{quote}
Path 1: Carbohydrate $\rightarrow$ Glucose (effect: +0.72)  

Path 2: Fiber $\rightarrow$ Protein (effect: -0.02) $\rightarrow$ Glucose (effect: -0.45)  

\end{quote}

\subsection*{Implementation of the Verification Step}
After retrieving candidate food items, the LLM is asked to verify whether the recommendation is consistent with the causal graph summary. 
The LLM must simulate the expected impact of the food’s nutrient profile on the outcome variable (glucose iAUC). 
If the recommendation does not align with the causal summary (e.g., suggesting a high-carbohydrate food when carbohydrate is identified as a strong glucose-raising factor), the LLM is instructed to reject the choice and propose an alternative.

The prompt for verification is:
\begin{quote}
Based on the causal summary, verify whether the candidate food item is appropriate for reducing glucose spikes.  
Food item: White rice (carbohydrates: 45g, fiber: 1g, protein: 3g, fat: 1g).  
Causal summary: Carbohydrates strongly increase glucose; fiber moderately decreases glucose; protein slightly decreases glucose.  
If the food contradicts the causal summary, explain why and suggest a better alternative.  
\end{quote}

The following steps are enforced:
\begin{itemize}
    \item The LLM must explicitly restate the causal summary before evaluating a food item.  
    \item Each candidate recommendation is passed through a “consistency check” where the nutrient profile is compared to the ranked causal paths.  
    \item If inconsistencies are found, the LLM is instructed to discard the recommendation and search for an alternative.  
    \item This process continues until a food item passes verification or a fallback recommendation (general dietary guideline) is returned.  
\end{itemize}

\subsection*{Implementation of the Path Ranking}

\textbf{Inputs.} Personal graph $G_i=(V_i,E_i,W)$, target $y$ (glucose iAUC at 30\,min / 1\,h / 2\,h), per-node usage statistic $R(v)$ (e.g., historical mean intake for nutrients or mean activity MET), and limits $(L_{\max}, P_{\max})$ for path length and number of scored paths.

\noindent\textbf{Scoring.} For each simple path $p = \{v_1,\ldots,y\}$ with length $\le L_{\max}$:
\[
S(p)= \sum_{(v_j,v_k)\in p} W_{jk}\, R(v_j)
\]
\noindent Aggregate node-level importance by summing path scores that include the node as a source:
\[
\mathrm{Imp}(v)= \sum_{p:\, v\in p} S(p)
\]
\noindent Rank paths by $S(p)$ and factors by $\mathrm{Imp}(v)$.
\begin{algorithm}[H]
\caption{Extract\_and\_Rank\_Causal\_Paths}
\label{alg:pathrank}
\begin{algorithmic}[1]
\STATE \textbf{Input:} $G_i=(V_i,E_i,W)$, target $y$, usage $R(\cdot)$, $L_{\max}$, $P_{\max}$
\STATE \textbf{Output:} Top paths $\mathcal{P}^\star$, factor ranking $\mathcal{F}^\star$
\STATE Initialize empty list $\mathcal{P}$ and map $\mathrm{Imp}[v]\leftarrow 0$ for $v\in V_i$
\FORALL{source node $s \in V_i$}
    \IF{$s = y$}
        \STATE \textbf{continue}
    \ENDIF
    \STATE Depth-limited DFS to enumerate simple paths to $y$:
    \STATE \textbf{function} DFS(current = $s$, path = [$s$])
    \IF{$|path| > L_{\max}$}
        \STATE \textbf{return}
    \ENDIF
    \IF{edge $(\mathrm{last}(path),y)\in E_i$}
        \STATE $p \leftarrow path \cup [y]$
        \STATE compute $S(p)=\sum_{(v_j,v_k)\in p} W_{jk}R(v_j)$
        \STATE append $(p,S(p))$ to $\mathcal{P}$
        \FORALL{$v \in p \setminus \{y\}$}
            \STATE $\mathrm{Imp}[v] \gets \mathrm{Imp}[v] + S(p)$
        \ENDFOR
    \ENDIF
    \FORALL{neighbor $u$ of $\mathrm{last}(path)$ with $(\mathrm{last}(path),u)\in E_i$ and $u \notin path$}
        \STATE DFS($u$, $path \cup [u]$)
    \ENDFOR
    \STATE DFS($s$, [$s$])
\ENDFOR
\STATE Sort $\mathcal{P}$ by $S(p)$ descending; keep top $P_{\max}$ as $\mathcal{P}^\star$
\STATE Sort nodes by $\mathrm{Imp}[v]$ descending to form $\mathcal{F}^\star$
\STATE \textbf{return} $\mathcal{P}^\star$, $\mathcal{F}^\star$
\end{algorithmic}
\end{algorithm}

\noindent\textbf{Hyperparameters.} We used $L_{\max}=3$ (nutrient/modulator $\rightarrow$ intermediate $\rightarrow y$) and $P_{\max}=5$. Results were stable for $L_{\max}\in\{2,3\}$ because of the size of the graph.






\subsection*{Sample Responses}
\paragraph*{Proposed System}
Question:
\{Suggest a breakfast that helps reduce postprandial glucose spikes.\}
Answer:
\{Your personal causal graph shows that carbohydrates strongly increase glucose (effect: +0.72). Fiber reduces glucose both directly and by enhancing the effect of protein, which together contribute to lower glucose excursions (combined effect: –0.45). Based on these pathways, a suitable breakfast is scrambled eggs with vegetables and whole grain toast. Scrambled eggs provide about 12 g protein and 9 g fat per serving, supporting glucose stability. Vegetables add fiber to reduce the glucose impact of carbohydrates, while whole grain toast contributes carbohydrates ($\approx$15 g) with additional fiber ($\approx$3 g) to buffer absorption. This combination balances the nutrient effects represented in your causal graph to reduce postprandial glucose spikes.\}

\paragraph*{ChatDiet}
Query: \{
Suggest a breakfast that helps reduce postprandial glucose spikes.\}
Answer: \{
To reduce postprandial glucose spikes, there are several nutrients that can be beneficial. Based on the provided list of food ingredients, I recommend incorporating the following food into your diet:
1. Scrambled Eggs with Whole Grain Toast: Eggs are a good source of protein and fat, which can slow down glucose absorption. Whole grain toast provides carbohydrates together with fiber, which helps to reduce sharp increases in blood glucose.
Aim to include meals with protein and fiber regularly, as this combination is known to support more stable glucose levels. To enhance the effectiveness of this recommendation, I suggest pairing it with regular physical activity.\}

\subsection{CGMacros Data Preparation}

\paragraph{Preprocessing}
We first convert timestamps to datetime objects and extract temporal features (date, time, hour, and minutes since last meal). Glucose signals are smoothed with a 5-point rolling mean and variability is assessed with a 5-point rolling standard deviation, reducing noise while preserving dynamic patterns. Macronutrient features (carbohydrates, protein, fat) are standardized, and total calories are computed using standard conversion factors (4 kcal/g for carbohydrates and protein, 9 kcal/g for fat).

\paragraph{iAUC Calculation}
Baseline glucose is defined as the average during the 120 minutes before each meal. Postprandial response is tracked over 120 minutes, with glucose values smoothed by a 5-point rolling mean. We compute glucose deltas relative to baseline, clipped at zero to capture only positive excursions. The iAUC is then calculated using the trapezoidal rule.

\subsection{Related Works and Research Gaps}

\begin{table*}[t]
\centering
\caption{Comparison of nutrition-oriented recommendation works highlighting the missing dimensions of structured, causal, and truly personalized reasoning.}
\label{tab:nutrition_gap}
\begin{tabular}{|l|p{3.0cm}|p{3.0cm}|p{3.0cm}|p{4.0cm}|}
\hline
\textbf{Paper} & \textbf{Personalization} & \textbf{Causal Reasoning} & \textbf{Adaptivity to Health}  & \textbf{Gaps} \\
\hline
De Hoogh et al. \cite{de2023design}& Conceptual: user preferences + general context & None (design framework only) & Not modeled & Provides design principles but no implementation of personalized causal inference \\
\hline

Nijman et al. \cite{nijman2007method} & None (population-level product scores) & None & Not applicable  & Nutrient-score system, not personalized or adaptive to individuals \\
\hline

Luo et al. \cite{luo2023reasoning} & General recommender systems & Explicit causal methods (PO, SCM, counterfactual) & Potential but not instantiated  & Signals promise of causal methods but no domain application to diet-health personalization \\
\hline
Yang et al. \cite{yang2024chatdiet} & Moderate: combines personal + population model via RAG & Implicit causal: retrieves causal triplets but fixed pattern & Limited adaptivity; retrieval-based only  & Improves personalization but not true dynamic causal reasoning; lacks graph traversal + counterfactual validation \\
\hline
Xue et al. \cite{xue2024dishagent} & Low: contextual personalization (smart dining) & None & Static (UX-oriented)  & Focused on experience, not metabolic personalization or health outcomes \\
\hline
Anjanamma et al. \cite{anjanamma2024personalized} & Moderate: behavior-based for children & None & Limited (behavioral adaptation only)  & Narrow population; lacks biomarker/casual reasoning; not generalizable to adults/clinical nutrition \\
\hline
\end{tabular}
\end{table*}

\end{document}